\documentclass[letterpaper,10pt,journal]{IEEEtran}
\usepackage[utf8]{inputenc}

\usepackage{graphicx}
\usepackage{xcolor}
\usepackage{siunitx}

\usepackage{enumitem}
\usepackage{amsmath}
\usepackage{amsfonts}
\usepackage{makecell}
\usepackage{booktabs}
\usepackage{multirow}
\usepackage{censor}
\PassOptionsToPackage{hyphens}{url} 
\usepackage{hyperref}
\hypersetup{hidelinks}

\usepackage[acronym]{glossaries}
\glsdisablehyper
\newacronym{ai}{AI}{artificial intelligence}
\newacronym{pci}{PCI}{percutaneous coronary intervention}
\newacronym{pvi}{PVI}{peripheral vascular intervention}
\newacronym{dsa}{DSA}{digital subtraction angiography}
\newacronym{ep}{EP}{electro physiological}
\newacronym{rcms}{RCMS}{robotic catheter manipulation system}
\newacronym{ip camera}{IP camera}{internet protocol camera}
\newacronym{rvir}{RVIR}{remote-controlled vascular interventional robot}
\newacronym{fbg}{FBG}{Fiber Bragg Grating}
\newacronym{smlr}{SMLR}{sparse multivariate linear regression}
\newacronym{vpn}{VPN}{virtual private network}
\newacronym{trl}{TRL}{Technology Readiness Levels}
\newacronym{fda}{FDA}{Food and Drug Administration}
\newacronym{isac}{ISAC}{Integrated Sensing and Communication}
\newacronym{mt}{MT}{mechanical thrombectomy}
\newacronym{vir2}{VIR-2}{vascular intervention robot 2}
\newacronym{ar}{AR}{augmented-reality}
\newacronym{or}{OR}{operating room}
\newacronym{ros2}{ROS 2}{Robot Operating System 2}
\newacronym{webrtc}{WebRTC}{Web Real-Time Communication}
\newacronym{5g}{5G}{fifth generation of cellular network technology}
\newacronym{wan}{WAN}{wide area network}
\newacronym{lan}{LAN}{local area network}
\newacronym{udp}{UDP}{user datagram protocol}
\newacronym{tcp}{TCP}{transmission control protocol}
\newacronym{rtt}{RTT}{round-trip time}
\newacronym{rica}{RICA}{right internal carotid artery}
\newacronym{rcca}{RCCA}{right common carotid artery}
\newacronym{mca}{MCA}{middle cerebral artery}
\newacronym{csma}{CSMA}{carrier sense multiple access}
\newacronym{p2p}{P2P}{point-to-point}
\newacronym{iqr}{IQR}{interquartile range}
\newacronym{sem}{SEM}{standard error of the mean}
\newacronym{geo}{GEO}{geostationary earth orbit}

\newacronym{ucid}{UCID}{user control input device}
\newacronym{rtp}{RTP}{Real-time Transport Protocol}
\newacronym{ssrc}{SSRC}{synchronization source identifier}
\newacronym{psnr}{PSNR}{peak signal-to-noise ratio}
\newacronym{ssim}{SSIM}{structural similarity index}
\newacronym{vmaf}{VMAF}{Video Multi-Method Assessment Fusion}
\newacronym{nasatlx}{NASA-TLX}{NASA Task Load Index}
\newacronym{reca}{RECA}{right external carotid artery}
\newacronym{ci}{CI}{confidence interval}

\title{Remote Teleoperation of Endovascular Robots: A Network Emulation Framework for Systematic Performance Evaluation
}

\author{
Xingyu Chen$^{1}$, Nikola Fischer$^{1}$, Harry Robertshaw$^{1}$, Weiyuan Deng$^{1}$, Christos Bergeles$^{1}$, and Thomas Booth $^{1,2}$
\thanks{
}
\thanks{
$^{1}$ School of Biomedical Engineering \& Imaging Sciences, King's College London, WC2R 2LS London, U.K.
$^{2}$ Department of Neuroradiology, King's College Hospital, SE5 9RS London, U.K.
}
}

\begin{document}
\bstctlcite{IEEEexample:BSTcontrol}

\maketitle

\begin{abstract}

Remote endovascular teleoperation requires network performance to be evaluated in relation to physical navigation and operator experience, while the effects of network degradation on the usability of remote endovascular systems remain insufficiently characterized from the perspective of clinical users. We developed a network emulation and benchmarking framework that combines an endovascular robot with data streaming. We conducted a user study involving eight clinicians from interventional neuroradiology and general interventional radiology and eight engineering students, who performed 480 navigation attempts across 15 network conditions. Communication performance and received video quality were assessed alongside navigation outcomes, confidence, workload and clinician rated acceptability. Under the most impaired combined condition, success was 78.1\%, compared with 93.8\% under the reference condition, and mean attempt time increased from 63.5 to 94.3\,s. At 5\% configured packet loss, 87.5\% of attempts succeeded, although all eight clinicians rated the condition unacceptable. Perceived video quality was associated with clinician confidence and acceptability, with mean within participant Spearman correlations of 0.72 and 0.78, respectively. Framework integration on a second robotic platform additionally supported successful remote navigation over a real network. Our framework provides a reusable basis that can help identify communication bottlenecks and assess their effects on navigation and clinician acceptance for endovascular intervention telesurgery tasks.

\end{abstract}

\glsresetall % The main text defines acronyms independently of the abstract.
\section{INTRODUCTION} \label{intro}

Endovascular interventions use guidewires and catheters to diagnose and treat diseases affecting the blood vessels through minimally invasive access with improved precision and reduced patient morbidity~\cite{miller_overview_2008}. Robotic manipulation can support controlled device navigation and separate the operator from the radiation environment, while providing a basis for remote intervention~\cite{patel_long_2019}. 

Despite progress in robotic endovascular interventions~\cite{crinnion2022robotics}, the process of remote teleoperation remains insufficiently addressed. Remote teleoperated endovascular robots could expand access to urgent procedures by enabling specialists to intervene from geographically distant locations~\cite{olson2026realtime}. This is particularly relevant to \gls{mt} for acute ischemic stroke, an effective treatment~\cite{goyal_endovascular_2016} whose functional benefit declines with each hour of delay~\cite{fransen_time_2016,saver_time_2016}. However, access to \gls{mt} remains unequitable, constrained by specialist availability and the geographical distribution of \gls{mt} capable centers~\cite{mcmeekin_estimating_2017}. In the United Kingdom, only 4.4\% of stroke patients receive \gls{mt} despite approximately 15\% being eligible~\cite{SSNAP2025StateNation}. Comparable disparities exist internationally: 8.4\% of ischemic stroke patients receive \gls{mt} in Germany~\cite{ungerer2024evolution}, 5.6\% across the United States~\cite{jafarli2025trends}, and 1.45\% in China~\cite{ye2022rates}. 

Patients presenting to local hospitals with imaging facilities but no \gls{mt} capability often require transfer to comprehensive stroke centers, adding treatment delays~\cite{sacks_multisociety_2018}. Remote robotic intervention could allow a distant specialist to perform \gls{mt} at a suitably equipped presenting hospital, potentially reducing transfer delays, mitigating local specialist shortages and improving resource allocation in time critical care~\cite{picozzi2023telemedicine}.

\begin{figure}[tb]
\centering
\includegraphics[width=1\linewidth]{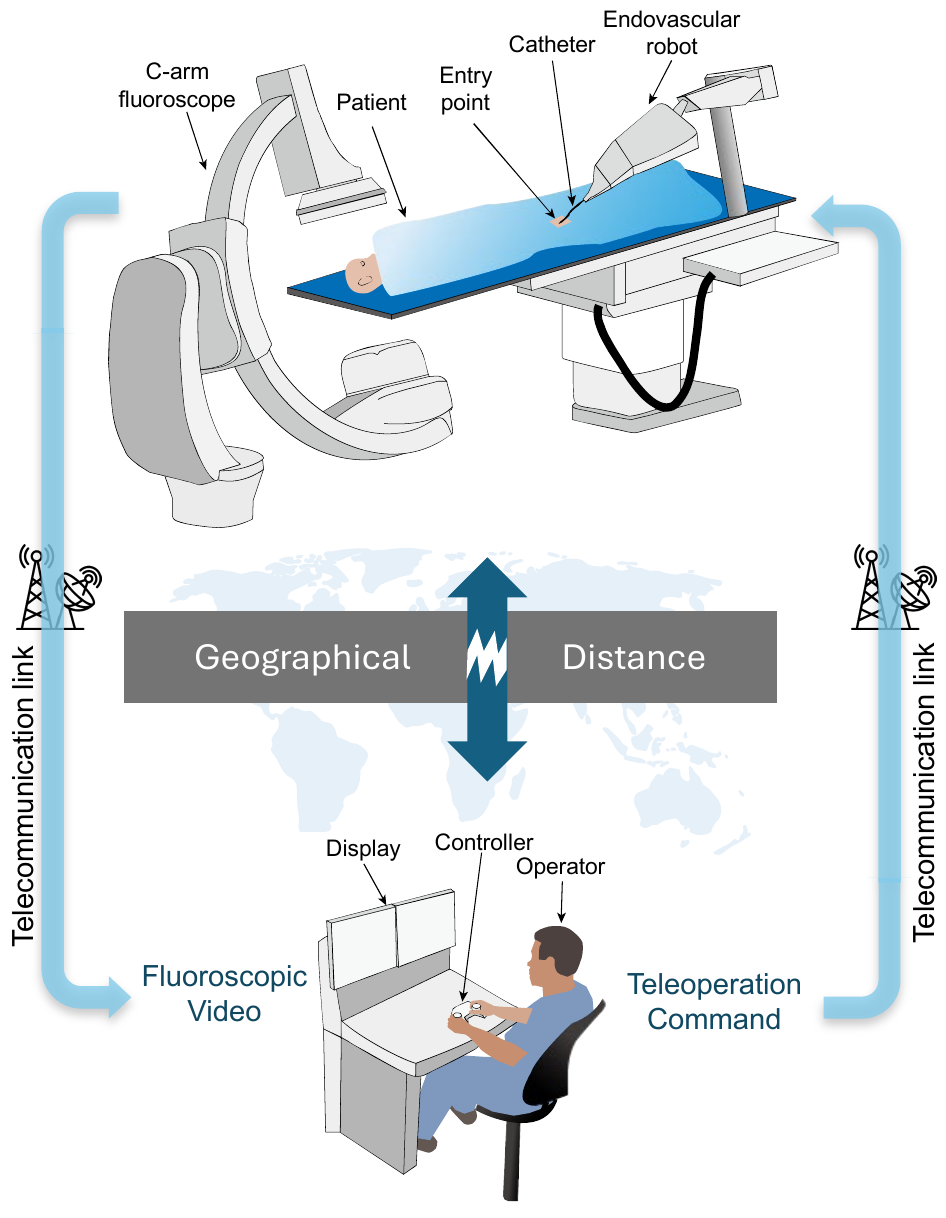}
\caption{Remote teleoperated endovascular intervention setup. The clinician operates a control console at a geographically separated site and transmits commands over the communication link to the patient-side endovascular robot in the operating room. Fluoroscopic video is streamed back to the operator to provide real-time visual feedback.}
\label{fig:tele_illu}
\end{figure}

Fig.~\ref{fig:tele_illu} illustrates a telerobotic configuration, in which the patient-side robot operates in the \gls{or} while the clinician manipulates the system in real time from a geographically separated control console via a telecommunication network. In systems that manipulate endovascular intervention instruments, the operator's inputs drive proximal translation and rotation, whereas navigation decisions depend on the distal tip's position and orientation within the vasculature. Instrument compliance and friction within the guidewire--catheter--vessel assembly complicate the relationship between proximal inputs and distal motion~\cite{kim2022telerobotic}. Operators use returned images to interpret motion and adjust subsequent commands~\cite{duckwiler2025remote}. Command transmission and visual feedback together close this human-in-the-loop process. Delayed commands affect motion timing, while delayed or degraded images alter the information available for correction. Network performance must therefore be evaluated in relation to the robot, its communication paths and the navigation task~\cite{chen2026remote}.

Clinical and preclinical studies have established several remote endovascular capabilities. Patel et al. demonstrated clinical remote \gls{pci} with on-site support~\cite{patel_long_2019}. Cao et al. developed a remote vascular robotic control system using \gls{webrtc} and \gls{5g}~\cite{cao2025webrtc}. Madder et al. performed transatlantic robotic coronary angiography in a phantom and compared on-site and remote catheter engagement across different Internet access modes~\cite{madder_transatlantic_2025}. In another study, long distance teleoperation has been reported as feasible at transcontinental scales, including separations up to \SI{9000}{\kilo\meter}~\cite{nelson_remote_2024}.  For neurovascular applications, Duckwiler et al. demonstrated integrated remote control with fluoroscopy and fluid management in a silicone phantom~\cite{duckwiler2025remote}. Olson et al. extended magnetic navigation to intercontinental phantom experiments and demonstrated clot aspiration in selected trials~\cite{olson2026realtime}.

Controlled impairment studies have begun to quantify how network conditions affect operation. Madder et al. introduced 0--1000\,ms of additional signal latency during coronary guidewire navigation in \textit{ex vivo} and \textit{in vivo} models, finding that completion time did not differ significantly across the tested range, whereas added delays of at least 400\,ms were reported as perceptible~\cite{madder_network_2020}. Zhang et al.'s NetFI study combined a physical surgeon console with a Davinc robotic simulation kits to evaluate delay, packet loss and communication interruptions while the returned video was not subjected to the injected impairments~\cite{zhang2026comprehensive}. Heemeyer et al. evaluated physical electromagnetic neurovascular navigation under combined latency, jitter, packet loss and bandwidth constraints with eight robotic engineers~\cite{heemeyer2026multidimensional}.

However, translating existing evidence into reproducible, system specific communication requirements for remote endovascular intervention remains challenging. A recent discussion paper from the U.S. \gls{fda} highlights degraded real-time imaging and asynchronous motion control and audio visual feedback as potential risks of remote teleoperation for endovascular stroke treatment~\cite{FDA2026RemoteTeleoperation}. Inconsistent reporting of network conditions and measurement endpoints limits comparison and replication across studies~\cite{chen2026remote}. Recent multi-parameter investigations have advanced the evaluation of physical endovascular navigation under controlled network impairments~\cite{heemeyer2026multidimensional}, but did not include clinician assessments across the tested network conditions. Nevertheless, configured network conditions alone do not describe the command and video performance experienced by the operator, while aggregate latency measures obscure differences between these channels. A remaining need is therefore to connect controlled network conditions, separately measured command and video transmission, and received visual quality with physical navigation outcomes and assessments by clinical users. Such an integrated evaluation is needed to establish how communication degradation affects task performance and operational acceptability, and to inform the design of networks, middleware and video pipelines for remote endovascular intervention. Such evaluation can guide system design and support the progression from demonstrations of remote feasibility towards reliable telesurgery.

In this work, we present a modular benchmarking framework that connects physical endovascular robots and their command and video traffic to configurable ns-3 network emulation~\cite{riley2010ns}, with a dedicated GStreamer pipeline for video feedback~\cite{GStreamerProject}. The main contributions of this study are summarized as follows: (i) a modular framework for systematically evaluating network conditions through separately measured command and video transmission and received visual quality. (ii) a user study of guidewire navigation in a vascular phantom involving 16 participants, comprising eight clinicians and eight engineering students, who completed 480 navigation attempts across 15 network conditions, combining representative network settings with targeted parameter sweeps. and (iii) implementations across distinct control architectures, including a \gls{ros2}-based system~\cite{Macenski2022ROS2}, and supplementary physical remote navigation over an actual network connection, further demonstrating the framework's applicability. To our knowledge, this is the first study to integrate systematic network performance evaluation of physical endovascular remote teleoperation with clinician informed assessments. The evaluation jointly characterizes command and video transmission, received visual quality, navigation outcomes and operator experience, providing a basis for interpreting technical degradation in the context of physical operation. The framework provides a common workflow for measuring communication performance alongside navigation outcomes and clinician assessments across robot implementations to identify system bottlenecks and inform communication requirements for future clinical application.

\section{Methodology}
\label{sec:methodology}

\subsection{Framework architecture and remote teleoperation implementation}
\label{sec:framework_overview}\label{sec:communication}
\begin{figure}[tb]
\centering
\includegraphics[width=1\linewidth]{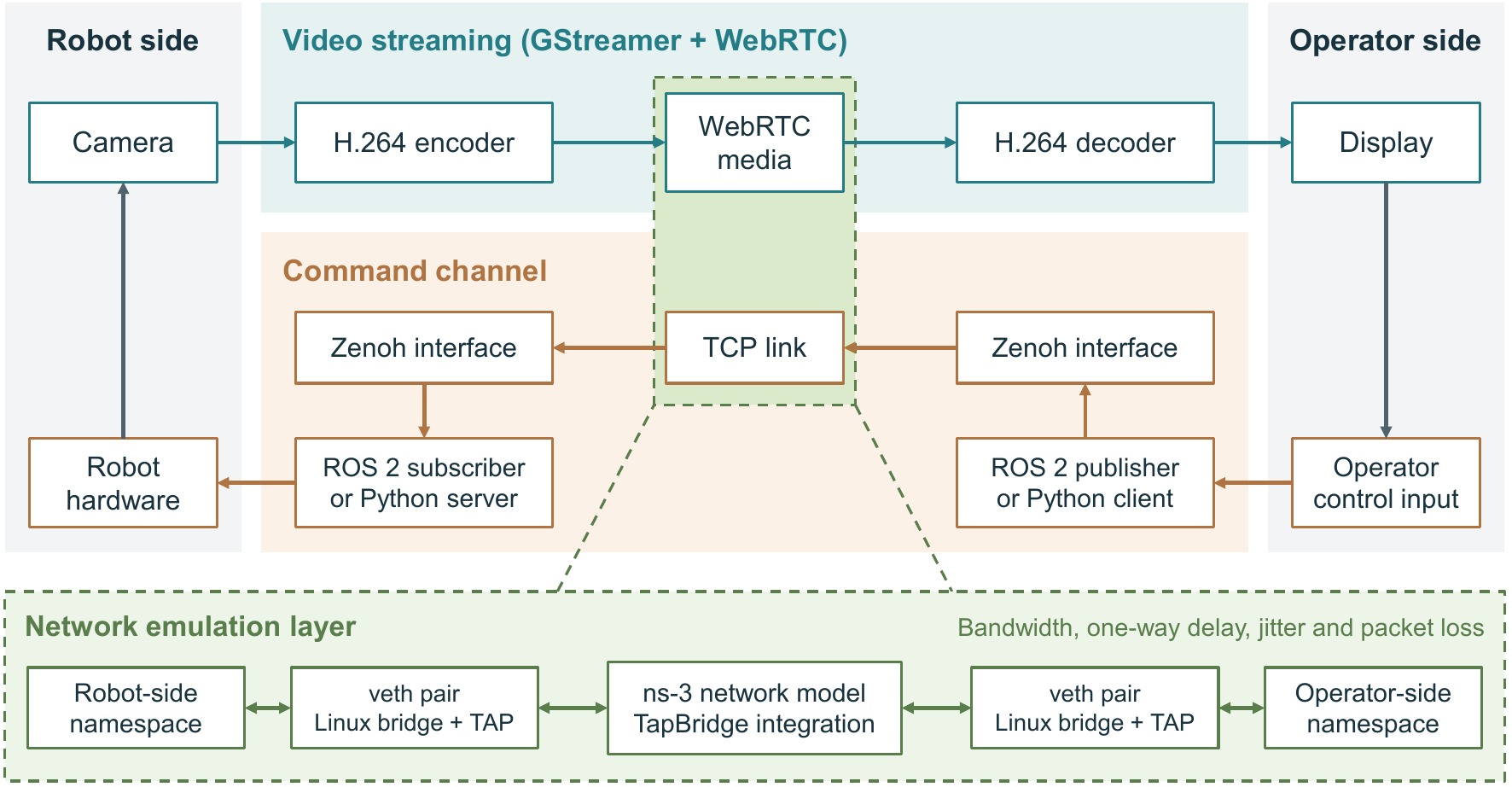}
\caption{Framework components and acquisition paths.}
\label{fig:framework}
\end{figure}

The benchmarking framework connects a physical endovascular intervention robot and an operator through command and video channels, with both channels passing through a configurable network-emulation layer (Fig.~\ref{fig:framework}). Its components comprise an operator-side \gls{ucid}, a command client, a robot-side controller, a camera observing the vascular phantom, a video encoder and receiver, and timestamped acquisition modules. The Python-controlled robotic platform was used for the primary phantom study while a second implementation connected the same communication and emulation components to a \gls{ros2} robot interface (Section~\ref{sec:ros_preliminary}).

Real application packets entered ns-3.46~\cite{riley2010ns} through Linux network namespaces, virtual interfaces and TAP bridges. Two \gls{csma} access networks were connected through a point-to-point bottleneck running with the ns-3 real-time simulator while the network conditions and emulation models are described in Section~\ref{sec:network_emulation}. Commands were transmitted through Zenoh over \gls{tcp}~\cite{ZenohWebsite} to a Python server interfacing with the robot motor controller at 20\,Hz rate. Visual feedback was provided by a camera viewing the phantom. The platform used H.264 over \gls{webrtc}, implemented in GStreamer~\cite{GStreamerProject}. The sender captures raw video through Video4Linux2, converts it to I420, overlays a frame identifier, and passes it through a bounded queue and the \texttt{x264enc} encoder. Parsed H.264 is packetized as \gls{rtp} and supplied to \gls{webrtc}. At the operator side, \gls{rtp} is depayloaded, H.264 is decoded, and frames are presented through an OpenCV display. The sender recording branches after encoding while the receiver recording received from the OpenCV display. The video configurations was set as $1920\times1080$ resolution at 30\,frames/s, a target encoder bitrate of 30,000\,kbit/s, a maximum key-frame interval of 60 frames, and zero B-frames. The \texttt{faster} preset and \texttt{zerolatency} tuning was selected for lower video latency with a one-buffer leaky queue and a 50-ms \gls{webrtc} receive buffer. These settings aim for consistent image format and encoding target while limiting accumulation of queued frames. Zero B-frames and low latency tuning are used to reduce encoder buffering~\cite{gstreamer_x264enc_docs}.

\subsection{Simulation network condition design and emulation models}
\label{sec:network_emulation}
The experiment combined two types of network conditions: representative combined conditions (C1--C4) with real world network references, and single parameter sweeps that varied one network factor while holding the remaining configuration fixed. Together, they support assessment across remote connection environments and of sensitivity to individual network factors. Table~\ref{tab:network_profiles} lists the network parameters including bandwidth, configured one-way delay, jitter (delay variation), and packet loss probability in each network conditions.

C1--C4 are approximations of real world network conditions. C1 represents regional terrestrial connectivity, with its 15\,ms nominal one-way delay consistent with published models of regional European \glspl{rtt}~\cite{martinez2024rtt}. C2 represents a longer distance international connectivity with measurements between London and US West (Silicon Valley) reported a 136\,ms \gls{rtt}, providing a reference for its delay scale~\cite{yao2026geolayer}. C3 represents degraded intercontinental connectivity between Europe and East Asia, with its 125\,ms one-way delay and 40\,Mbps bandwidth informed by London to Beijing measurements~\cite{yao2026geolayer}. C4 represents connectivity through a \gls{geo} satellite link with approximately 250\,ms propagation scale based on ITU-R M.2460-0~\cite{itu_m2460_2019}. This condition illustrates connectivity to remote sites with limited terrestrial access, such as rural communities in parts of Africa served by satellite backhaul~\cite{spengler2020geostationary}.

The parameter combinations in Table~\ref{tab:network_profiles} use these network scales as references, with progressively reduced bandwidth and increased jitter and packet loss. Published London to Zhangjiakou measurements showed increased \gls{rtt} and packet loss during throughput degradation, supporting evaluation under combined impairments~\cite{zhu2020transnational}. The packet loss was then set according to ITU-T Y.1541 which provides a $10^{-3}$ packet-loss objective as background for the low loss settings~\cite{itu_y1541_2011},while the higher loss levels in C3 and C4 extend the evaluation to degraded connectivity.

Each sweep changed only one of configured parameter while fixing the others for comparable results. Relative to C1, the delay sweep used delay of 100, 200 and 300\,ms (DL1--DL3). Earlier telesurgery experiments motivate examining delays on this scale~\cite{nankaku2022delay}. The jitter sweep used jitter equal to 20, 50 and 80\,ms (JI1--JI3), referenced to DL2, the delay-sweep condition with delay of 200\,ms and jitter of 1\,ms. The loss sweep used levels of 1\%, 3\% and 5\% (PL1--PL3), and the bandwidth sweep used 40, 30 and 25\,Mbps (BA1--BA3), both using C1 for comparison. The initial design comprised 16 distinct network conditions including four representative conditions, three additional delay conditions, three jitter conditions, three loss conditions and three bandwidth conditions as shown in Table~\ref{tab:network_profiles}. BA3 at 25\,Mbps was not used in the user study experiment after preliminary testing because video freezing prevented task completion. The analyzed user study therefore comprised 15 network conditions.

\begin{table}[tb]
\centering\small
\caption{Network conditions used in the experiments.}
\label{tab:network_profiles}\label{tab:network_sweeps}
\setlength{\tabcolsep}{2pt}
\begin{tabular*}{\columnwidth}{@{\extracolsep{\fill}}lcccc@{}}\toprule
\shortstack[l]{Condition /\\sweep} & \shortstack{Bandwidth\\(Mbps)} & \shortstack{Delay\\(ms)} & \shortstack{Jitter\\(ms)} & \shortstack{Loss\\(\%)}\\\midrule
\multicolumn{5}{@{}l}{\textit{Representative conditions}}\\
C1 & 60 & 15 & 1 & 0.01\\
C2 & 45 & 60 & 5 & 0.10\\
C3 & 40 & 125 & 10 & 0.50\\
C4 & 30 & 250 & 20 & 1.00\\\midrule
\multicolumn{5}{@{}l}{\textit{Single-parameter sweeps}}\\
Delay & 60 & \textbf{100, 200, 300} & 1 & 0.01\\
Jitter & 60 & 200 & \textbf{20, 50, 80} & 0.01\\
Loss & 60 & 15 & 1 & \textbf{1, 3, 5}\\
Bandwidth & \textbf{40, 30} & 15 & 1 & 0.01\\\bottomrule
\end{tabular*}
\par\smallskip
\begin{minipage}{\columnwidth}\footnotesize
Delay is the configured nominal one-way link delay. Bold lists give additional settings applied separately. Reused references: C1 for delay, loss and bandwidth and DL2 for jitter.
\par\smallskip
\end{minipage}
\end{table}

Delay variation was generated independently in each transmission direction by linearly interpolating stochastic targets updated every $h$ milliseconds, introducing temporal correlation. Let $D$ denote the nominal one-way delay and $J$ the configured jitter parameter. Independent draws $Z_{k,q}\sim\mathcal{N}(0,1)$ are indexed by target update $k$ and direction $q$. Starting from $a_{0,q}=\max\{0,D+\sqrt{3/2}\,JZ_{0,q}\}$, the targets and interpolated delay are defined as
\begin{equation}
\begin{aligned}
 a_{k+1,q}&=\max\{0,D+\sqrt{3/2}\,JZ_{k+1,q},a_{k,q}-0.95h\},\\
 d_q(t)&=a_{k,q}+\frac{t-kh}{h}(a_{k+1,q}-a_{k,q}),
\end{aligned}
\label{eq:delay_model}
\end{equation}
where \(t\) is the simulation time, \(k=0,1,2,\ldots\) indexes target updates, and $kh\leq t<(k+1)h$ denotes the interval between successive updates. All times are expressed in milliseconds. 
The constraints prevent negative delays and limit their downward slope to preserve packet arrival ordering. The factor $\sqrt{3/2}$ compensates for  interpolation-induced variance reduction before constraints are applied. The configured delay $D$ and jitter $J$ define the intended network condition while the realized packet delay statistics may slightly differ from these nominal settings.

Packet loss was simulated at each transmission direction using the ns-3 packet level \texttt{RateErrorModel}~\cite{ns3_error_model_docs}. Independent uniform random streams were used for each transmission direction. For packet $i$ in direction $q$, the uniform draw $U_{i,q}$ and receive error indicator $X_{i,q}$ were defined as
\begin{equation}
 U_{i,q}\sim\mathcal U(0,1),\qquad X_{i,q}=\mathbf 1\{U_{i,q}<P\},
 \label{eq:packet_loss_model}
\end{equation}
where $X_{i,q}=1$ denotes a receive-error drop and $P$ is the configured packet loss probability. Thus a 1\% setting means $p=0.01$ at the receiving endpoint in each direction.

Bandwidth limitation was implemented through the point-to-point link data rate $B$~\cite{ns3_point_to_point_docs}. For packet $i$ in direction $q$, let $L_i$ be its size in bytes as used by that device and $t_i$ its transmission start time. With $B$ in bit/s and all time terms converted to seconds, the local link time before any arrival order adjustment was
\begin{equation}
 T_{\mathrm{link},i}=Q_i+\frac{8L_i}{B}+d_q(t_i),
 \label{eq:serialization}
\end{equation}
where $8L_i/B$ is the serialization time, $d_q(t_i)$ is the network delay evaluated at transmission start and $Q_i$ its queue residence time. Each transmission direction had its own drop-tail queue, allowing queueing delay and overflow loss.

\subsection{Communication measurements and video quality assessment}
\label{sec:measurement}
Command and video timing were quantified as one-way latency using the host system clock shared by the endpoints in the single-host namespace experiments. For sample $k$, integer Unix timestamps in nanoseconds were subtracted before conversion to milliseconds:
\begin{equation}
\begin{aligned}
 \tau^c_k&=\bigl(t^c_{\mathrm{recv},k}-t^c_{\mathrm{send},k}\bigr)/10^6,\\
 \tau^v_k&=\bigl(t^v_{\mathrm{recv},k}-t^v_{\mathrm{stamp},k}\bigr)/10^6.
\end{aligned}
\label{eq:one_way_delay}
\end{equation}
Here $t^c_{\mathrm{send},k}$ and $t^c_{\mathrm{recv},k}$ denote client probe transmission and server probe receipt, respectively. For video, $t^v_{\mathrm{stamp},k}$ denotes timestamp insertion into the \gls{rtp} extension after encoding, and $t^v_{\mathrm{recv},k}$ is recorded at the receiver \gls{webrtc} \gls{rtp} output, before depayloading and decoding. The command measure excludes motor execution. The video measure includes queueing, packet transmission and propagation, \gls{webrtc} jitter-buffer waiting and intervening software processing while it does not include camera capture, encoding, decoding and display latency. These application level one-way latencies are distinct from the configured ns-3 propagation delay. Echo-based \gls{rtt} was used for timestamp consistency checks.

Receiver \gls{rtp} loss was derived from consecutive \gls{webrtc} packet counters, separately for each \gls{ssrc}. Over valid within-\gls{ssrc} intervals $\mathcal V$, the recording-level loss percentage was
\begin{equation}
 \widehat p_{\mathrm{RTP}}=100\,
 \frac{\sum_{k\in\mathcal V}\Delta N_{\mathrm{lost},k}}
 {\sum_{k\in\mathcal V}(\Delta N_{\mathrm{received},k}+\Delta N_{\mathrm{lost},k})}.
 \label{eq:rtp_loss}
\end{equation}
Here $\Delta N_{\mathrm{received},k}$ and $\Delta N_{\mathrm{lost},k}$ denote validated received and lost counter changes, pooled across eligible intervals and \glspl{ssrc}, measuring \gls{rtp} accounting. Receiver interarrival jitter was sampled from \gls{webrtc} inbound \gls{rtp} statistics for each \gls{ssrc} and converted from seconds to milliseconds~\cite{w3c_webrtc_stats}, following the standard RFC~3550 estimator~\cite{schulzrinne2003rtp}.

Video quality was assessed against the sender-side encoded recording, using the corresponding receiver recording. After alignment to a common recorded frame identifier and timestamp, subsequent decoded frames were paired by sequential input index and the longer tail was truncated. This pairing can retain temporal mismatches caused by dropped or duplicated frames. \Gls{psnr} was computed over validated unmasked pixels, excluding the marker region. \gls{ssim}~\cite{wang2004ssim} and  \gls{vmaf}~\cite{netflix_vmaf} were computed over full frames and averaged across paired frames. These comparisons characterize received video fidelity, including temporal mismatch and recording effects relative to an encoded reference.

\subsection{Primary phantom study and outcome measures}
\label{sec:primary_study}

\begin{figure*}[!t]
\centering
\includegraphics[width=\linewidth]{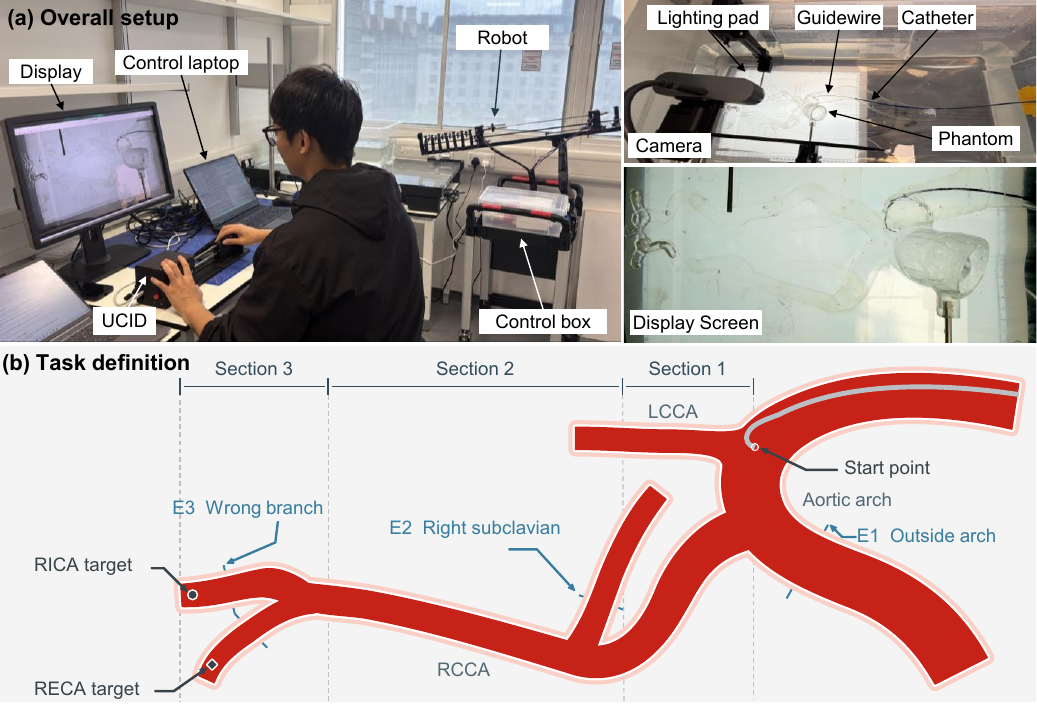}
\caption{User study setup and task illustration. (a) Operator console with user control input device (UCID), robot, guidewire/catheter assembly, vascular phantom, overhead camera and backlighting, with the camera image displayed to the operator. (b) Navigation task diagram with standardized start, RICA and RECA targets, and three navigation sections. E1--E3 denote the illustrated error locations labeled outside arch, right subclavian and wrong branch.}
\label{fig:experiment_task}\label{fig:primary_setup}
\end{figure*}

\paragraph{Experimental setup and navigation task}
The experimental platform used a custom-built endovascular robot~\cite{sadati2025autonomous} and device-mimicking \gls{ucid}~\cite{jackson2026manual}. These have been validated in combination during previous user studies~\cite{fischer2026copilot}. Sliding and rotating the \gls{ucid} commanded axial translation and rotation of the guidewire through a Python controller. Operators used direct manual control and conventional camera feedback, with no shared autonomy~\cite{fischer2026copilot}. The endovascular instruments comprised a Guidewire (180 cm, 0.035 in, Terumo Corporation, Tokyo, Japan) and two supporting catheters including a Penumbra catheter (100 cm, 0.044 in, Penumbra, Inc., Alameda, CA, USA) and Neuron MAX 088 (Penumbra, Inc., Alameda, CA, USA). As in previous autonomous navigation studies~\cite{robertshaw2026autonomous}, transparent vascular phantom was 3D-printed in Clear V4 resin (Formlabs, Inc., Somerville, MA, USA) placed inside a box full of glycerin for transparency, and a light pad (TOHETO, Ganzhou, China) was placed beneath for backlighting with 2000-lx setting. A full HD webcam (Logitech, Lausanne, Switzerland) above the phantom supplied the operator's visual feedback as shown in Fig.~\ref{fig:experiment_task}a, while image acquisition and transmission settings are given in Section~\ref{sec:communication}.

The task comprised navigation from a standardized guidewire starting position in the aortic arch to the \gls{rica} or \gls{reca}. Each network condition contained one attempt per target, with a 150\,s limit per attempt. The robot and endovascular devices were reset to the same starting position before the next navigation attempt. Fig.~\ref{fig:experiment_task}b shows the three navigation sections including initial arch navigation, advancement along the \gls{rcca}, and the final approach to the assigned target.

\paragraph{Participants and experimental procedure}
The analyzed cohort comprised eight clinicians and eight novices giving $16\times15\times2=480$ navigation attempts. The study used a within participant design, with each participant completing the two target task under all 15 network conditions. Clinicians included consultants and specialty trainees in interventional neuroradiology or general interventional radiology, with reported endovascular experience of 1--18 years (median 7.5 years). The novices were engineering students with no prior clinical experience in endovascular interventions.

Two minutes of free familiarization and one practice attempt to each target are set before main navigation to minimized the affection of learning curve. Participants were assigned to eight predefined sequences, each used once in each participant group. Four sequences placed the representative condition block first and four placed the sweep block first. Representative condition order and single parameter groups order were balanced across their four positions. After completing both targets, with the instruments reset between attempts, participants were asked to completed the corresponding questionnaire. The detailed condition sequences and target orders can be found in Supplementary S1.

\paragraph{Navigation outcome measurements}
For each attempt, we recorded target success, three section times and error counts~\cite{robertshaw2026position} in three categories: outside the aortic arch, subclavian and wrong branch. The protocol defined termination at target entry or timeout after 150\,s. For successful attempts, total navigation time was the sum of the three recorded section times. The anatomical setting and task completion criteria are shown in Fig.~\ref{fig:experiment_task}b. The target was reached when the guidewire tip was 10\,mm inside the assigned branch, without buckling that prevented further advancement. The three error categories corresponded to E1, E2 and E3, respectively, and required guidewire tip advancement of 10\,mm into the corresponding incorrect branch. A further error was counted each time the guidewire tip was withdrawn from an incorrect branch and subsequently re-entered it to this depth. Section times accumulated according to the section occupied by the guidewire tip. Timing resumed for a previously visited section whenever the tip returned to it, and switched again when the tip entered another section.

\paragraph{Subjective outcomes from questionnaire}
Questionnaires were completed once per network condition after both target attempts. Condition questionnaires assessed overall video quality on a five-category scale (bad to excellent), perceived delay and confidence on 1--7 scales, and clinician rated acceptability on a 1--7 agreement scale. Higher delay scores indicate more perceived delay while higher quality, confidence and acceptability scores indicate more favorable ratings. The representative condition questionnaire additionally assessed visibility, artifacts, control response and six \gls{nasatlx} dimensions~\cite{hart1988nasatlx} on 0--100 scales. Single parameter groups used the shorter form. Workload dimensions were reported individually, without a composite \gls{nasatlx} score. Detailed questionnaire wording, response anchors and scoring are given in Supplementary S1.

\subsection{Data processing and statistical analysis}
\label{sec:statistical_analysis}
Analyses of the primary phantom study used participants as the unit of condition comparison and resampling, with two navigation targets treated as different measurements within each participant. Successful attempts retained observed completion time while failed navigation attempts were assigned the 150\,s limit.
This endpoint reflects both completion speed and task success. Success rates used all included attempts and error means were averaged the two targets within participant. For technical metrics, valid recording summaries were averaged equally within participant and network condition, then across participants while latency summaries used recording medians.

Time differences between network conditions were calculated for matched references and averaged within participant. Sweep references were C1 for delay, packet loss and bandwidth, and DL2 for jitter. Exploratory comparisons used two sided exact paired signed rank sign permutation tests with Holm correction across all pairwise comparisons within each condition set. Time and error summaries used 95\% percentile \glspl{ci} from 10,000 participant bootstrap resamples. Trial order analyses compared early and late exposure groups within condition. Valid condition ratings were summarized by medians and \glspl{iqr}, and \gls{nasatlx} dimensions by means and standard deviations. Acceptability summaries included only clinicians. Within each participant, spearman correlations were calculated across common valid network conditions between subjective video quality or perceived delay ratings and confidence, acceptability or navigation outcomes.

\subsection{Application to a second robotic platform and remote deployment}
\label{sec:ros_preliminary}

To assess framework integration and remote deployment on a second robot, we included a second robot platform implementation as a separate application study. The robot used two alternating cartridges, stepper motor actuation and pneumatic gripping, with \gls{ros2} nodes providing the robot facing interface. The design of the robot can be found in~\cite{fischer2026compact}. It shared the same communication and simulation framework. The operator console issued motion commands while receiving camera feedback. 

Two clinical trainees and one engineering student performed three \gls{rcca} to \gls{rica} navigation for each condition with a total of 54 attempts. The task limit was 5\,min, and conditions followed the fixed order Local, sim1, sim2, sim3, sim4 and Remote. These conditions were independent of the primary study's C1--C4 and their definitions are given in Supplementary S3. Remote connected workstations in separate buildings approximately 200\,m apart through operator side Ethernet, robot side institutional Wi-Fi and a ZeroTier \gls{vpn}. \textit{sim4} was parameterized using measurements of the network connection used for the Remote condition. It provided a measurement informed emulated condition for comparison of communication and navigation outcomes with the real deployment. Outcomes were target reach success and completion time. Command and video measurements were reported as condition level medians of estimated one-way transmission latency with detail of measurement method can be found in supplementary S3. Results were summarized descriptively, separately from the primary study and the fixed condition order precludes separating network effects from practice.

\section{Results}
\label{sec:results}

The primary evaluation examines navigation outcomes and operator ratings under controlled network conditions. The evaluation involved 16 participants, comprising eight clinicians and eight novice participants. The questionnaire contains  240 participant to condition ratings, 120 clinician acceptability ratings, and workload ratings from all 16 participants under C1--C4. Another evaluation on the \gls{ros2}-based robotics platform is reported separately as evidence of implementation feasibility. 

\begin{figure}[!tb]
\centering
\includegraphics[width=\linewidth]{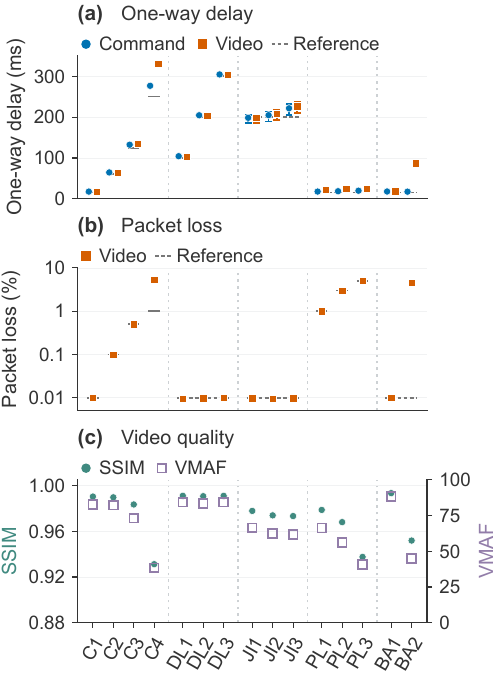}
\caption{Measured communication performance and video quality across 15 network conditions, comprising representative conditions (C1–C4) and delay (DL1–DL3), jitter (JI1–JI3), packet loss (PL1–PL3), and bandwidth (BA1–BA2) sweeps. (a) One-way delay for command and video traffic. (b) Video RTP packet loss on a logarithmic scale. (c) Video quality quantified by the structural similarity index (SSIM) and Video Multi-method Assessment Fusion (VMAF). Error bars in (a,b) are 95\% participant-bootstrap confidence intervals while (c) shows means only.}
\label{fig:measured_performance}
\end{figure}

\subsection{Measured communication performance and video quality}
\label{sec:results_technical}

Mean command and video one-way delays increased from 17.6 and 17.0\,ms in C1 to 276.7 and 331.5\,ms in C4, respectively (Fig.~\ref{fig:measured_performance}a). Across DL1 to DL3, command one-way delay rose from 104.2 to 304.5\,ms and video one-way delay from 103.0 to 303.5\,ms. Both followed the nominal one-way delay references of 100, 200 and 300\,ms. The reference values represent the configured one-way link delay, rather than the full instrumented delay. At BA2, video one-way delay reached 87.0\,ms compared with 17.4\,ms for commands.

Receiver-observed \gls{rtp} packet loss remained near 0.01\% in C1 and the delay and jitter sweeps as shown in Fig.~\ref{fig:measured_performance}b. Across PL1 to PL3, mean observed loss increased from 1.00\% to 3.01\% and 5.01\%, closely matching the configured percentages. Reducing bandwidth from 40\,Mbps (BA1) to 30\,Mbps (BA2) increased observed loss from 0.010\% to 4.47\%, a difference of 4.46 percentage points at the same configured loss of 0.01\% ($P_{adj}<0.001$). C4 had the highest mean observed loss, 5.37\%, compared with its configured 1\%. This condition combined several network impairments.

Relative to the sender-side encoded reference, mean \gls{ssim} and \gls{vmaf} were 0.990 and 82.8 in C1 (Fig.~\ref{fig:measured_performance}c). Delay sweep means remained close to C1, whereas jitter sweep means were lower than their DL2 reference. At JI3, the mean differences from DL2 were $-0.0174$ for SSIM ($P_{adj}<0.001$) and $-21.7$ for VMAF ($P_{adj}<0.001$). Both metrics declined across the packet loss sweep and were lower in BA2 than BA1 ($P_{adj}<0.001$). C4 had the lowest means, 0.931 for \gls{ssim} and 38.4 for \gls{vmaf}. BA1 gave the highest mean for each metric.

\subsection{Navigation performance under representative network conditions}
\label{sec:results_representative}

\begin{figure*}[!t]
\centering
\includegraphics[width=\linewidth]{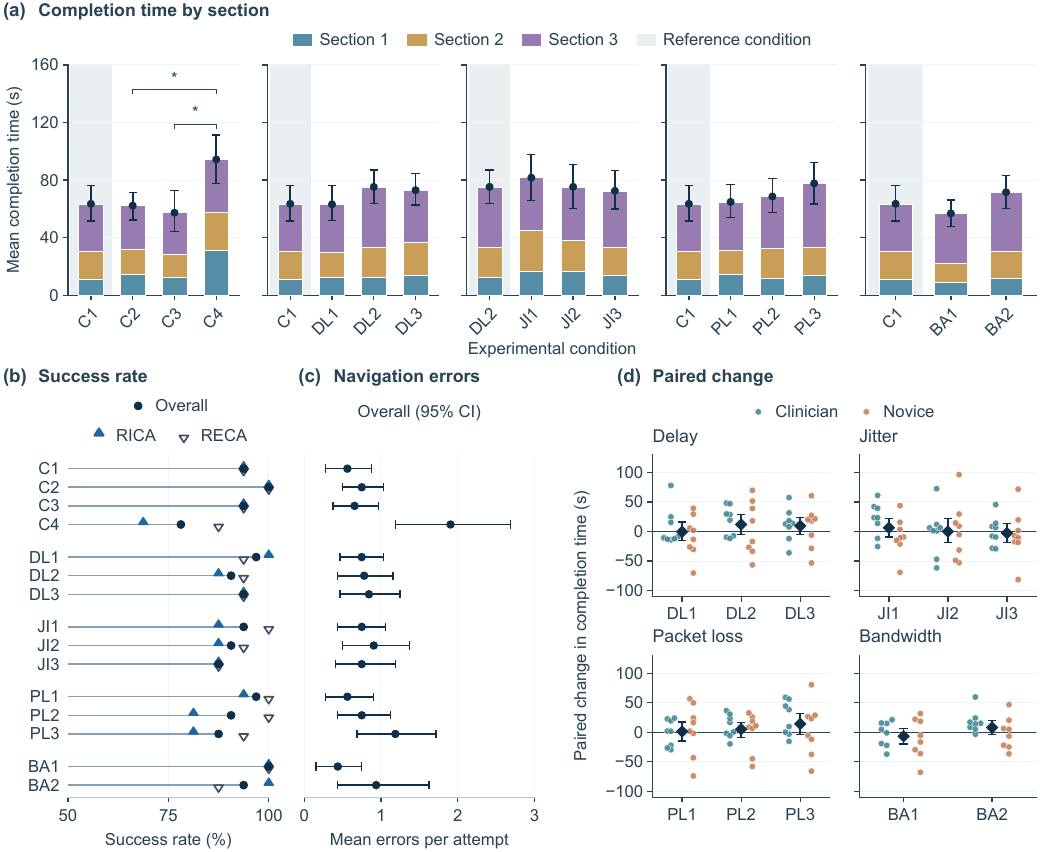}
\caption{Navigation success, error counts and completion time across network conditions. (a) Mean times for Sections 1--3, stacked to show total completion time, targets are averaged within participants. (b) Target specific success rates for right internal carotid artery (RICA) and right external carotid artery (RECA), C1--C4 denote the four representative network conditions. (c) Mean errors per attempt, summing outside aortic arch, subclavian, wrong branch errors. Targets are averaged within each of all participants. (d) Within participant time differences relative to C1 for delay, packet loss and bandwidth, and DL2 for jitter. Differences between matched targets are averaged within each participant. Pale shading marks the reference condition in each group. Repeated reference bars reuse the same observations. Failed attempts are assigned 150\,s and their Section 3 value is the remainder after the recorded Sections 1 and 2, not a measured completion duration. Brackets mark significant pairwise contrasts: $^{*}P<0.05$ after Holm correction and two sided exact paired signed-rank sign-permutation tests.}
\label{fig:success_completion}
\end{figure*}

Completion time comparisons included all attempts, with failures assigned 150\,s. Paired differences were averaged across the two targets within each participant. Across the full dataset, 444/480 attempts were successful (92.5\%). In C1 and C3, 30/32 attempts succeeded (93.8\%), and all 32 succeeded in C2. C4 had the lowest observed success rate at 25/32 attempts (78.1\%). Success in C4 was lower for the \gls{rica} than for the \gls{reca} target (11/16 versus 14/16, Fig.~\ref{fig:success_completion}b). Supplementary S2 gives the outcomes by participant and target.

Mean navigation times were 94.3\,s in C4 and 63.5\,s in C1 (Fig.~\ref{fig:success_completion}a). The C4--C1 difference was 30.8\,s (48.5\%). The exact paired test gave an unadjusted $P=0.039$, but the difference did not remain significant after Holm correction ($P_{\mathrm{adj}}=0.155$). C4 shows significant increase from C2 by 32.1\,s ($P_{\mathrm{adj}}=0.026$) and from C3 by 37.0\,s ($P_{\mathrm{adj}}=0.016$).

Section 1, initial arch navigation, accounted for most of the C4--C1 mean increase. Its mean rose from 10.8 to 31.6\,s, contributing 20.8\,s (67.5\%) of the 30.8\,s total difference while sections 2 and 3 increased by 6.8 and 3.3\,s.  The failed attempts Section 3 values are assigned remainders, not measured completion durations.  Overall, the time and success patterns were not monotonic across C1--C4, although C4 had the longest mean time and lowest observed success.

Mean errors per attempt ranged from 0.56 to 0.75 in C1--C3 and reached 1.91 in C4 (Fig.~\ref{fig:success_completion}c). C4 had approximately 2.5--3.4 times as many errors per attempt as the other three conditions. Outside aortic arch errors (E1) accounted for the largest increase, averaging 1.22 per attempt in C4 compared with 0.13--0.31 in C1--C3. These errors constituted 63.9\% of all errors recorded in C4. Error counts by category and condition are given in Supplementary S2. Overall, C4 had the highest mean error count, alongside the longest mean attempt time and the lowest success rate.

\subsection{Navigation sensitivity to individual network impairments}
\label{sec:results_parameter_sensitivity}

The parameter sweeps showed different patterns across delay, jitter, packet loss and bandwidth as shown in Fig.~\ref{fig:success_completion}. C1 was the reference for delay, packet loss and bandwidth  while jitter comparisons used DL2 at a fixed 200\,ms configured one-way delay.

\paragraph{Delay}
Success varied little as configured one-way delay increased from 15\,ms in C1 to 300\,ms in DL3. At both settings, participants reached the target in 30/32 attempts. Mean navigation time was close to C1 at DL1. The paired time differences from C1 were $+11.8$\,s at DL2 and $+9.5$\,s at DL3. These time differences were not monotonic, varied across participants and did not reach statistical significance (Fig.~\ref{fig:success_completion}d). And section 3 accounted for 9.1\,s of the 11.8\,s increase in mean navigation time from C1 to DL2. Mean errors per attempt increased slightly from 0.56 in C1 to 0.84 in DL3 (Fig.~\ref{fig:success_completion}c).

\paragraph{Jitter}
At a fixed 200\,ms delay, mean time was longest at 20\,ms jitter (JI1). At 80\,ms (JI3), it fell below the mean for DL2, the 1\,ms jitter reference (Fig.~\ref{fig:success_completion}a). Across the three increased-jitter settings, failures increased from two to four. The largest mean difference from DL2 was 6.4\,s at JI1 ($-9.6$ to $21.5$\,s). None of the paired completion-time differences from DL2 reached statistical significance. The mean difference in completion time is mainly contributed by Section 2, which contribute to the 93.4\% of the reduction. Mean errors per attempt showed no monotonic pattern.

\paragraph{Packet loss}
Mean time increased across the packet loss sweep and was 14.2\,s longer at 5\% configured loss (PL3) than at 0.01\% (C1) (Fig.~\ref{fig:success_completion}d). Success was highest at PL1 (31/32) and fell to 28/32 (87.5\%) at PL3. Mean errors per attempt increased from 0.56 in C1 to 1.19 in PL3. Outside aortic arch counts increased from 4 to 19, and wrong-branch counts from 8 to 15. Paired mean time differences relative to C1 were $+1.3$, $+5.1$ and $+14.2$\,s (Fig.~\ref{fig:success_completion}d), while the increase was more consistently monotonic among clinicians.

\paragraph{Bandwidth}
Mean time at 40\,Mbps (BA1) was shorter than at 60\,Mbps (C1), and all 32 BA1 attempts succeeded (Fig.~\ref{fig:success_completion}a,b). Two target pairs that failed in C1 but succeeded in BA1 accounted for 6.3\,s of the 6.6\,s difference in mean time. The lower mean therefore does not establish a benefit of reducing bandwidth. Errors increased from 0.44 to 0.94 per attempt as bandwidth decreased from BA1 to BA2. The paired BA2--BA1 time difference was $+14.7$\,s, with a more apparent increase among clinicians. 

\paragraph{Navigation performance by trial order}
\label{sec:results_learning_curve}
Navigation times were compared between early and late exposure groups within each network condition using the trial-order definition in Supplementary S2. Point estimates favored shorter times with late exposure for C1--C4 and longer times for DL1--DL3, with varying directions across the remaining conditions. Late-minus-early differences were $-32.7$\,s  for C3 and $+34.9$\,s ($18.4$\,s to $51.1$\,s) for DL2. Error-count differences varied in direction, only C3 and DL3 had intervals entirely below zero. In C3, shorter navigation time coincided with fewer errors: the late-minus-early difference was $-0.81$ errors per attempt.

\subsection{Operator experience}
\label{sec:results_operator_experience}

\begin{figure*}[!tb]
\centering
\includegraphics[width=\linewidth]{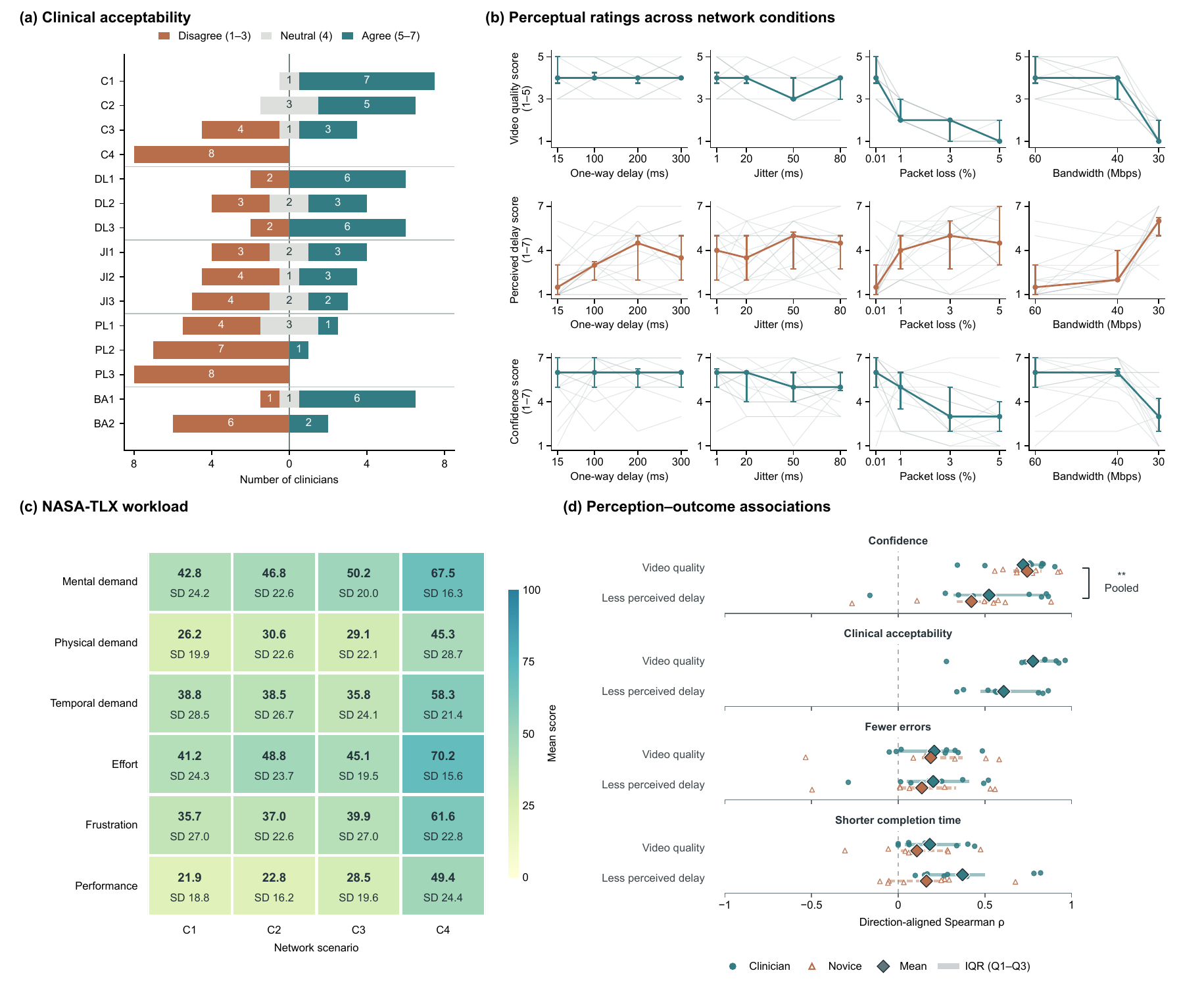}
\caption{Operator experience across network conditions and associations with task outcomes. (a) Acceptability ratings from eight clinicians: disagreement (1--3), neutral (4), and agreement (5--7), for the phantom task. (b) Video quality, perceived delay and confidence. Colored points show medians, error bars interquartile ranges (IQRs), and pale lines individual ratings. Video quality was rated on a five-point scale, whereas perceived delay and confidence were rated on seven-point scales. (c) NASA-TLX Workload scores across C1--C4, shown as means and standard deviations. Higher value indicate greater workload. (d) Direction-aligned within-participant Spearman correlations across conditions. Teal circles and solid bars denote clinicians; orange triangles and dashed bars denote  novices. Diamonds show cohort means, and bars show interquartile ranges. Correlation signs were aligned so that positive values indicate better perceived video quality or less perceived delay accompanying more  favorable outcomes. Acceptability includes eight clinicians only while other outcomes include all participants. The bracket denotes the pooled confidence correlations ($^{**}P<0.01$).}
\label{fig:questionnaire}
\label{fig:primary_operator_ratings}
\label{fig:primary_joint_assessment}
\end{figure*}

\paragraph{Perceived video quality and delay}
Median video quality remained close to 4 on the five-point scale throughout the delay and jitter sweeps (Fig.~\ref{fig:questionnaire}b). Across the packet loss settings, median video quality scores were 2, 2 and 1, with most participants reporting progressively poorer quality. Reducing bandwidth from 40 to 30\,Mbps (BA1 to BA2) lowered median video quality from 4 to 1. Perceived delay scores varied considerably among participants in the delay, jitter and packet loss sweeps, whereas the bandwidth sweep showed a more consistent monotonic pattern. Median perceived delay was higher at each of DL1--DL3 than in C1. The median perceived-delay rating remained around 4.5 across the jitter settings.

\paragraph{Confidence and workload}
Confidence remained relatively high throughout the delay sweep, with a median of 6 on the seven-point scale at all three settings. Median confidence decreased slightly from 6 in JI1 to 5 in both JI2 and JI3. The decline was larger in the packet loss sweep, from 5 in PL1 to 3 in both PL2 and PL3. From BA1 to BA2, median confidence fell from 6 to 3.
All six \gls{nasatlx} dimensions reached their highest mean scores in C4 (Fig.~\ref{fig:questionnaire}c). Physical demand, temporal demand, effort and performance scores remained similar across C1--C3. Mental demand and frustration increased progressively from C1 to C3 and rose further in C4. Effort and frustration showed the largest increases from C3 to C4, rising by 25.1 and 21.7 points, respectively.

\paragraph{Clinician-rated acceptability}
Responses on the seven-point scale were grouped for visualization as disagreement (1--3), neutral (4) and agreement (5--7). Across C1--C4, clinician-rated acceptability decreased monotonically (Fig.~\ref{fig:questionnaire}a). Most clinicians agreed that the conditions were acceptable for the task in C1 (7/8). Half disagreed in C3 (4/8), and none of the clinicians agreed in C4 (0/8).
Agreement showed no monotonic trend across the three delay levels, with fewer clinicians agreeing at DL2 than at either DL1 or DL3. Across the three jitter levels, agreement remained low at two or three of eight clinicians. Disagreement increased with packet loss, from 4/8 clinicians at PL1 to 7/8 at PL2. At PL3, all eight clinicians disagreed that the condition was acceptable for the task, although 28/32 navigation attempts succeeded. Two participants reported that they could hear the motors during the same room experiment and used this sound to help interpret the timing of the robot's response and where sluggishness might arise.

\subsection{Associations among technical, perceptual, and task outcomes}
\label{sec:results_joint_assessment}

\paragraph{Technical measurements and operator perceptions}
Across DL1--DL3, mean command and video one-way delays increased monotonically, whereas median perceived-delay ratings peaked at DL2 and fell at DL3 (Figs.~\ref{fig:measured_performance}a and~\ref{fig:questionnaire}b). For video quality, the condition means for \gls{ssim} and \gls{vmaf} and the median participant ratings showed similar declines across the packet loss sweep and from BA1 to BA2 (Figs.~\ref{fig:measured_performance}c and~\ref{fig:questionnaire}b).

\paragraph{Within participant analysis}
We examined how measured communication performance and received video fidelity corresponded to operator perceptions, and how those perceptions related to confidence, clinician-rated acceptability and navigation outcomes. The associations shown in Fig.~\ref{fig:questionnaire}d were quantified using within-participant Spearman rank correlations. Perceived video quality was more consistently associated with confidence than was less perceived delay. Means of the Spearman correlations were 0.72 for clinicians and 0.74 for novices, compared with 0.52 and 0.42 for less perceived delay. The video quality correlation was larger in 14 of 16 participants ($P=0.0021$). Among clinicians, acceptability had a mean correlation of 0.78 for video quality and 0.61 for less perceived delay.

Associations with navigation outcomes were numerically smaller than those with confidence and acceptability. For both cohorts, the mean correlations with fewer errors were similar for video quality and less perceived delay. Mean correlations with shorter completion time were 0.18 and 0.11 for video quality, and 0.37 and 0.16 for less perceived delay.  Among clinicians, less perceived delay had a higher mean correlation with shorter completion time than did video quality (0.37 versus 0.18).

\subsection{Cross platform application and real network remote teleoperation}
\label{sec:results_cross_architecture}

On the second robotic platform, three operators navigated the physical guidewire across local, emulated and remote network conditions (Table~\ref{tab:ros2_execution}). Overall, 53 of the 54 attempts reached the target. The sole unsuccessful attempt occurred in sim3 and was attributed to a robot malfunction. Across the six conditions, median completion times among successful attempts ranged from 92.4 to 125.7~s. The lowest and highest condition medians occurred in sim2 and Local, respectively.

\begin{table}[!htb]
\centering
\caption{Navigation outcomes and estimated one-way transmission latency on the ROS 2-based robotic platform.}
\label{tab:ros2_execution}
{\color{black}
\small
\setlength{\tabcolsep}{4pt}
\renewcommand{\arraystretch}{1.15}
\begin{tabular*}{\columnwidth}{@{\extracolsep{\fill}}lcccc@{}}
\toprule
\multirow{2}{*}{Condition} &
\multirow{2}{*}{\shortstack{Reached\\($n/N$)}} &
\multirow{2}{*}{\shortstack{Time (s)\\Median}} &
\multicolumn{2}{c}{\shortstack{Median\\Latency (ms)}} \\
\cmidrule(lr){4-5}
& & & Command & Video \\
\midrule
Local & 9/9 & 125.7 & --- & --- \\
sim1 & 9/9 & 99.2 & 14.0 & 21.1 \\
sim2 & 9/9 & 92.4 & 44.0 & 50.4 \\
sim3 & 8/9 & 106.8 & 392.2 & 522.5 \\
sim4 & 9/9 & 96.4 & 8.0 & 12.9 \\
Remote & 9/9 & 110.5 & 8.6 & 16.8 \\
\bottomrule
\end{tabular*}
\par\vspace{3pt}
\parbox{\columnwidth}{\scriptsize
Reached denotes successful attempts/all attempts. Time is the median completion time among successful attempts within each condition. Command and video values are medians of estimated one-way transmission latency. Dashes indicate unreported values. Detailed measurement method and condition description can be found in Supplementary S3.

}
}
\end{table}

Across conditions with reported communication values, median estimated one-way transmission latency ranged from 8.0 to 392.2\,ms for commands and from 12.9 to 522.5\,ms for video. The highest medians occurred in sim3, at 392.2\,ms for commands and 522.5\,ms for video. Local latency was unreported and was excluded from these ranges. As described in Section~\ref{sec:ros_preliminary}, sim4 was parameterized using measurements of the network connection used for Remote. Median estimated one-way transmission latencies were 8.0 and 8.6~ms for commands and 12.9 and 16.8\,ms for video in sim4 and Remote. Both conditions achieved 9/9 successful navigations, with median completion times of 96.4 and 110.5\,s. Their latency medians were in a similar range, whereas the Remote completion-time median was 14.1\,s higher.

\section{Discussion}

This study establishes an adaptable framework for controlled network evaluation with physical robots and real application traffic. The main finding is that nominal network severity, delivered feedback and navigation usability are related but distinct. By linking controlled impairments to measured communication performance, received video quality, navigation outcomes and operator experience, the framework reveals where these relationships diverge. Condition C4 showed the poorest feedback and navigation outcomes, while successful target attainment could coexist with low confidence and clinician disagreement about acceptability. These findings support evaluating the complete operating context when defining communication requirements for endovascular remote teleoperation.

\subsection{Network Impairments and Usable Visual Feedback}

Condition C4 illustrated why single parameter severity cannot predict delivered feedback. Its configured loss of 1\% was below PL2/PL3 (3\%/5\%), and its 250\,ms one-way delay was below DL3 (300\,ms). Nevertheless, C4 had the highest observed video \gls{rtp} loss (5.37\%, versus 5.01\% in PL3), lowest video fidelity and poorest principal navigation outcomes. Command/video latencies were 276.7/331.5~ms in C4, compared with 304.5/303.5~ms in DL3. Thus, a condition with lower nominal delay produced slower video delivery while commands arrived sooner. This supports joint evaluation of impairments, consistent with previous multi-parameter study~\cite{heemeyer2026multidimensional}.

Visual feedback was especially closely associated with the operator's confidence and judgment. Mean within participant video quality to confidence correlations were 0.72 for clinicians and 0.74 for novices, compared with 0.52 and 0.42 for less perceived delay. Clinician acceptability correlated more strongly with video quality (0.78) than with less perceived delay (0.61). Navigation requires reliable judgments of guidewire tip position, orientation and movement, as well as timely images. Objective fidelity scores, subjective quality ratings and this task relevant information are complementary measure. Consoli et al. likewise found that artificially imposed video latency and reduced image refresh rates were associated with lower operator acceptability and dangerous uncontrolled movements in simulated neurovascular procedures~\cite{consoli2025image}.

Bandwidth requirements depended on the video configuration. At BA2, 30\,Mbps capacity matched the 30,000\,kbit/s H.264 target bitrate before transport overhead. Observed \gls{rtp} loss reached 4.47\%, and video latency rose to 87.0\,ms despite command latency of 17.4\,ms. BA3 at 25\,Mbps caused severe freezing during preliminary testing and was excluded from the user study. Limited capacity headroom makes queueing, overflow and buffering plausible explanations for this channel divergence. The practical implication is to characterize application traffic and its fluctuations alongside bottleneck capacity. The command and video channels constitute a minimal functional communication configuration for the navigation task tested here. A complete remote procedure may also require \gls{or} scene video, bidirectional audiovisual communication with the bedside team, robot and equipment status, force or torque information, and additional clinical images or camera views. These channels have different data volumes, timing and reliability requirements. Communication provisioning should therefore account for the complete workflow, including concurrent traffic and timing priorities. By enabling controlled evaluation of application traffic under different network conditions, the framework provides a basis for investigating these requirements as additional workflow channels are introduced.

\subsection{Task-Specific Performance and Operator Interpretation}

The concentration of deterioration in particular task components gives more useful design information than total time alone. Section~1 contributed 20.8\,s, or 67.5\%, of the descriptive 30.8\,s C4--C1 increase, while outside aortic arch errors (E1) constituted 63.9\% of C4 errors. Initial arch navigation requires tip localization, continuous trajectory correction and timely confirmation of the response to keep the instrument on the intended vessel path. Less usable feedback may disrupt these linked judgments. Successful navigation did not ensure clinician acceptance. In PL3, 28/32 attempts succeeded, yet all eight clinicians reported that the condition was not acceptable for the phantom task. Reduced confidence, higher error counts, increased \gls{nasatlx} effort and frustration in C4, and completion time changes together describe burdens that success alone misses. Operators may preserve target attainment by changing input pacing, observing for longer, using task familiarity or making additional corrections.  A previous study similarly reported successful porcine endovascular navigation without significant navigation time differences across added latencies, despite increased perceived lag and procedural impact~\cite{legeza_impact_2022}.

Perceived response delay can also reflect the mechanical transmission between proximal actuation and distal motion. There are previous studies characterized friction and backlash in guidewire--sheath catheter actuation~\cite{kesner2011position} and documented translational and rotational hysteresis, including friction, mechanical play and stored rotational elastic energy~\cite{alahmad2023force} in a robotic catheterization system. These mechanical effects differ from a fixed communication delay while actuator motion need not produce an immediately corresponding tip displacement. The displayed response also includes the video pipeline. Consequently, clinicians observing delayed tip motion may have difficulty determining whether the limiting stage is communication, robot actuation or instrument transmission. And two participants reported using audible motor activity to interpret response timing and where sluggishness might arise. Because the primary experiment took place in the same room, sound provided an additional cue about proximal actuator activity beyond the displayed procedure view. Such a cue may indicate that execution has begun, but cannot establish that the distal guidewire tip has moved. 

\subsection{Implications for Feedback Design and Benchmarking}

Additional state feedback may help operators distinguish command transmission, proximal actuation and distal instrument response. The reports of motor sounds assisting response interpretation suggest that such supplementary cues may already influence operator perception. Explicit proximal and distal motion or force feedback could therefore be tested when visual responses are delayed or ambiguous. Although previous research found no significant benefit from haptic feedback during \textit{in silico} endovascular navigation tasks~\cite{jackson2023comparative}, its potential role in explaining the execution process during remote teleoperation remains to be investigated.

Our framework can be used to test changes to communication or feedback under the same network conditions. Transmission latency and received video quality can then be compared alongside navigation performance, operator confidence, workload and acceptability. Reporting the robot interface, task and measurement endpoints makes these comparisons interpretable and reproducible.

The applicability of the framework beyond the primary experimental setup was further examined using a second \gls{ros2} robotic platform and a real network connection. All nine remote attempts reached the target over the Ethernet, institutional Wi-Fi and \gls{vpn} connection. Parameterizing sim4 from measurements of that connection demonstrates how deployment observations can inform controlled condition design. This is evidence of framework integration and deployment feasibility under the tested circumstances. 

\subsection{Limitations and Future Directions}

The primary study involved eight clinicians and eight engineering students navigating to two targets in one fixed phantom. Task familiarity and trial-order associated variation constrain transfer beyond this setting. Additional anatomies and navigation tasks would test the persistence of the observed patterns. Clinical fluoroscopy and multiple views should also be evaluated because the camera images used here differ in contrast, texture and motion, affecting both transmission demands and task-relevant visibility.

The timing measures end at command receipt and receiver \gls{rtp} output, excluding robot execution and several video stages. Synchronized measurements of operator input, command arrival, actuation, distal guidewire motion and display response would separate communication effects from mechanical response. Remote experiments that control auditory access and vary explicit state, force or other multimodal feedback are needed to test the proposed interpretive benefit alongside navigation outcomes.

Network coverage was limited to the selected delay variation, independent loss models, capacities and application settings. Dynamic congestion, burst loss, interruptions and concurrent workflow channels would test resource competition and temporal reliability more directly. Further real-network deployments could inform richer controlled conditions and assess their correspondence with observed application behavior. The second-platform study used three operators, a fixed condition order and a separate protocol; aligned protocols and measurement endpoints would be needed for quantitative cross-platform comparison.

\section{Conclusion}
This work presents a network emulation and benchmarking framework for evaluating physical endovascular robots under controlled communication conditions. The primary endovascular study demonstrates that transmission latency, received video quality, task completion and perceived acceptability provide complementary information while successful navigation alone does not capture the operating burden. Application to a second robotic platform and navigation over a real network connection demonstrate reuse of the framework, while simulation conditions illustrates the use of deployment measurements to construct a controlled condition. This workflow supports controlled comparisons of network configurations and provides a basis for evaluating alternative operator feedback designs for a specified robotic platform and navigation task.

\bibliographystyle{IEEEtran}
\bibliography{IEEE}

\end{document}